\documentclass[runningheads]{llncs}

\usepackage{epsfig}
\usepackage{graphicx}
\usepackage{amsmath}
\usepackage{amssymb}

\usepackage{booktabs}
\usepackage{multirow}
\usepackage{subfig}
\usepackage{wrapfig}
\usepackage{enumitem}
\DeclareCaptionLabelFormat{andtable}{#1~#2  \&  \tablename~\thetable}
\usepackage[font=small,skip=3pt]{caption}

\usepackage[pagebackref=true,breaklinks=true,letterpaper=true,colorlinks,bookmarks=false]{hyperref}

\begin{document}


\title{Scientific Discovery by Generating Counterfactuals using Image Translation}

\author{Arunachalam Narayanaswamy$^{\star 1}$ 
\and
Subhashini Venugopalan\thanks{equal contribution}$^{1}$  
\and
Dale R. Webster$^{2}$ 
\and
Lily Peng$^{2}$ 
\and
Greg S. Corrado$^{2}$ 
\and
Paisan Ruamviboonsuk$^{3}$  
\and
Pinal Bavishi$^{2}$ 
\and
Rory Sayres$^{2}$
\and
Abigail Huang$^{2}$
\and
Siva Balasubramanian$^{2}$\thanks{via Adecco Staffing, California.}
\and
Michael Brenner$^{1}$ 
\and
Philip C. Nelson$^{1}$ 
\and
Avinash V. Varadarajan$^{2}$ 
}
\institute{
$^{1}$Google Research, 
$^{2}$Google Health, 
$^{3}$Rajavithi Hospital
}

\maketitle
\vspace{-8pt}
\begin{abstract}
Model explanation techniques play a critical role in understanding the source of a model's performance and making its decisions transparent. Here we investigate if explanation techniques can also be used as a mechanism for scientific discovery. We make three contributions: first, we propose a framework to convert predictions from explanation techniques to a mechanism of discovery. Second, we show how generative models in combination with black-box predictors can be used to generate hypotheses (without human priors) that can be critically examined. Third, with these techniques we study classification models for retinal images predicting Diabetic Macular Edema (DME), where recent work~\cite{varadarajan2018predicting} showed that a CNN trained on these images is likely learning novel features in the image. 
We demonstrate that the proposed framework is able to explain the underlying scientific mechanism, thus bridging the gap between the model's performance and human understanding.

\end{abstract}

\section{Introduction}
Visual recognition models are receiving increased attention in the medical domain \cite{varadarajan2018predicting,gulshan2016development}, supplementing/complementing physicians and enabling screening at a larger scale, e.g. in drug studies \cite{kaggle2019}. While the application of deep neural net based models in the medical field has been growing, 
it is also important to develop techniques to make their decisions transparent.
Further, these models provide a rich surface for scientific discovery. 
Recent works~\cite{poplin2018prediction,pmlr-v97-miller19a} show that neural net models can make novel predictions previously unbeknownst to humans. However for such works, explanation techniques based on saliency are insufficient. Specifically, saliency maps only show spatial support i.e.~``where" the model looks, but do not explain ``what" is different (or ``why"). %

There are a few critical differences between explanation methods for natural images and medical data. Firstly, while humans have a direct intuition for objects in natural images, only trained professionals typically read medical data with acceptable accuracy~\cite{krause2018grader}. Further, %
such medical training and knowledge is accrued over time driven by advances in basic science. %
However, such knowledge is rarely injected when developing deep models for medical data.
Many of these novel predictions are based on raw data and direct outcomes, not necessarily human labels. Such end-to-end models show improved performance while simultaneously reducing human involvement in knowledge distillation. Thus, there is a compelling need to develop explanation methods that can look at data without fully-known human interpretable annotations  %
and generate hypotheses both to validate models and improve human understanding.
\begin{figure}%
\vspace{-6pt}
    \centering
    \includegraphics[width=0.7\textwidth]{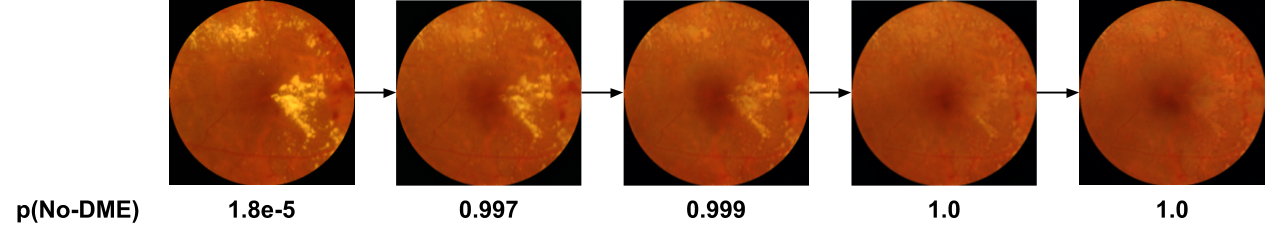}
    \caption{\small{An example of a transformation from left to right by successively applying our explanation technique based on unpaired image to image translation to modify the source image with diabetic macular edema (DME) (\textit{leftmost column}) to no-DME (\textit{rightmost column}). Probability of no-DME from an independent prediction model are presented below. Our method %
    accurately shows ``what" changes affect prediction.
    }}%
    \label{fig:intro_figure}%
\end{figure}

In this work we propose to use image-to-image translation in combination with black-box prediction models to achieve this goal. Specifically, we first use existing techniques to identify salient regions relevant to the prediction task. We then develop image translation models that can accurately modify image regions of a source class to that of a target class to show what about the region influences prediction. %
Further, we are able to amplify the modifications to enhance human interpretability.
Finally, based on the transformations observed, we identify a minimal set of hand-engineered features which when trained using a linear SVM achieves comparable performance to that of the CNN trained on the raw images. 
To demonstrate our approach we look at the Diabetic Macular Edema (DME) prediction models from ~\cite{varadarajan2018predicting}. In their work, the same CNN architecture shows drastically different performance when trained using labels from 2 different sources. We reproduce that and explain how.
Our contributions are:
\begin{itemize}
    \item A framework to convert predictions from black-box models to a mechanism of discovery.
    \item A demonstration of how image translation models in combination with black-box predictors can be used to generate hypotheses worth examining.
    \item A set of hand-engineered features, identified from the generated hypotheses, that account for the performance of a Diabetic Macular Edema classifier.
\end{itemize}

\noindent\textbf{Related Work.} The past few years have seen several methods for explaining classification models. While some focus on visualizing the concepts learned by the convolutional filters~\cite{mahendran2015understanding,bau2017network,mordvintsev2015deepdream},
a predominant number focus on generating saliency masks. %
These are 
more relevant to our work.
Of the saliency %
methods, some are backpropagation based \cite{selvaraju2017grad,springenberg2014striving,sundararajan2017axiomatic,fong2017interpretable,kapishnikov2019xrai} and 
examine the model's gradients to produce a mask.
Some others are pertubation based \cite{Zeiler_2014,fong2017interpretable,fong2019understanding,ribeiro2016should,petsiuk2018rise,smilkov2017smoothgrad} treating the model as a black-box and observe its outputs while perturbing the inputs. %
In both cases the methods generate a heatmap to estimate \emph{where} the prediction model is looking and reveal spatial support. %
While saliency maps are helpful to validate models, they don't provide the complete picture. In particular, they don't reveal \emph{what} about these regions is different between two classes. %
Our method bridges this gap by generating counterfactuals akin to \cite{dhurandhar2018explanations,samangouei2018explaingan,liu2019generative,joshi2019towards,chang2018explaining,singla2019explanation}.
Specifically, we use image translations, to show sparse changes on the original image, to reveal subtle differences between classes. We also amplify these changes to enhance human interpretability while simultaneously producing realistic images that doctors can analyze (Figures \ref{fig:intro_figure}, \ref{fig:cyclegan_qualitative}).

\section{Approach}\label{sec:methods}
Given a deep prediction model that can distinguish between two image classes, we wish to discover features that are relevant for classification.
One objective way to accomplish this is to identify a distilled set of features computed from the image that carry nearly the same predictive power as the model. %
In this work, instead of purely human driven hypotheses to discover features, our framework uses model explanation techniques as a tool for generating these hypotheses.

\textbf{1. Input ablation to evaluate the importance of known regions.}
With human annotations of known landmarks/regions in medical images, it is possible to evaluate how relevant these are for prediction. %
Individually ablating each known landmark from the dataset, retraining from scratch, and evaluating the prediction model %
can help understand whether there are compact known salient regions in the data that influence the model. 

\textbf{2. Visual saliency to validate the model.} Saliency techniques~\cite{selvaraju2017grad,springenberg2014striving,sundararajan2017axiomatic} allow for arbitrary spatial support. They can be used to validate the model based on known features or regions identified by the ablation experiments in the previous step. In natural images, with known object classes, the regions can be human interpretable but that may not always be the case for medical images.

\textbf{3. Image-to-image translation to highlight ``what" is different.}
Next, to show ``what" is different between the 2 classes %
we propose unpaired image to image translation using Generative Adversarial Networks (GANs)~\cite{goodfellow2014generative}.
Here, we use a slightly modified version of CycleGAN~\cite{zhu2017unpaired}. 
Let us refer to the two image classes as $X$ and $Y$.
The basic idea behind unpaired image translation is to learn 2 functions: $g: X \rightarrow Y$, and $f: Y \rightarrow X$, for translating images from one class to the other as determined by the GAN discriminator. CycleGAN~\cite{zhu2017unpaired} incorporates a cycle consistency loss in addition to the two GAN losses. This ensures that $f \circ g$ and $g \circ f$ ($\circ$ refers to composition) are close to identity functions i.e. $f(g(x)) \approx x$ and $g(f(y)) \approx y$. The total loss being optimized can be written as 
\begin{align*}
    L_{total} &= L_{GAN} (G, D_y, X, Y) + L_{GAN}(F, D_x, X, Y) + \lambda L_{Cycle} \\
    L_{Cycle} &= \mathbb{E}_{x \sim X}(\lVert f(g(x) - x \rVert_1) + \mathbb{E}_{y \sim Y}(\lVert g(f(y) - y \rVert_1)
\end{align*}
$G, F$ refer to the generators, and $D_x, D_y$ the discriminators within the GAN functions $g$ and $f$ respectively. $L_{GAN}$ refers to the GAN loss function (we use a wasserstein loss~\cite{gulrajani2017improved}). 
Next, we modified the basic CycleGAN model architecture (above) to retain high frequency information during this translation:
\begin{itemize}
    \item we have a $1 \times 1$ convolutional path from the input to the output, and
    \item we also model the functions as residual networks.
\end{itemize}
Such a construction allows the model to copy pixels from input to output if they aren't relevant to the classification. This allowed our model to quickly learn the identity function and then optimize for cycle consistency to highlight subtle differences between classes in the counterfactual.

To verify that the image transformation to the target class is successful we need a prediction model independent of the CycleGAN discriminator. It is necessary that the CycleGAN has no interaction with the prediction model to ensure that it cannot encode any class information directly into the translation functions (see \cite{chu2017cyclegan}).
We use a CNN, which we call $M$, trained on the same set of images ($X$ and $Y$), completely independent of the CycleGAN model. In order to verify that our CycleGAN model succeeded, we evaluate area under the curve (AUC) of the translated and original images under model $M$.

\textbf{4. Amplify differences from image transformation.} Next we amplify the differences in the counterfactual %
by applying 
the translation function successively i.e $g(x), g(g(x)), \cdots g^4(x)$, (similarly for $f$). To evaluate this quantitatively we compute AUCs for all examples in $(x, y), (g(x), f(y)), \cdots (g^4(x), f^4(y))$ using the prediction model $M$. If the the model $M$ has an AUC of $0.85$ on original images $(x, y)$ with successive applications of $f$ and $g$, we would expect this to drop to $0.15$ if the image translation is able to perfectly map the two class of images. We also evaluate the images qualitatively (as in Figures~\ref{fig:intro_figure} and ~\ref{fig:cyclegan_qualitative}).

\textbf{5. Identify and evaluate hand-engineered features.}
Recall that our goal is to discover and explain what about the data leads to the predictive power of the classification model. To this end, based on the visualizations generated from the image translation model, we designed specific hand-engineered features observed in the transformations. We then evaluate them using 2 simple classifiers, a linear SVM with $l1$ regularization, and an MLP with 1 hidden layer. We compare the performance of these classifiers on the hand-engineered features with that of the CNN model ($M$) on a held out evaluation set.

\section{Diagnosing DME - 2 sources of labels}
Diabetic Macular Edema(DME)~\cite{lee2015epidemiology} %
, a leading cause of preventable blindness, %
is characterized by retinal thickening of the macula.
The gold standard for DME diagnosis involves the evaluation of a 3D scan of the retina, based on Optical Coherence Tomography(OCT), by a trained expert.
However, due to costs and availability constraints, screening for DME generally happens by examining a Color Fundus Photo (CFP), which is a 2D image of the retina. Human experts evaluating a CFP look for the presence of lipid deposits called hard exudates (HE) which is a proxy feature for retinal thickening~\cite{harding1995sensitivity}. This method of screening for DME through HE is known to have both poor precision and recall when compared to the gold standard diagnosis~\cite{wang2016comparison,mackenzie2011sdoct}.

\begin{figure}
    \centering
    \subfloat[\small{no DME}]{\includegraphics[width=0.18\textwidth]{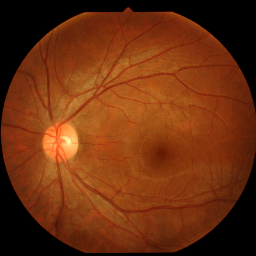}}
    \subfloat[\small{no DME}]{\includegraphics[width=0.18\textwidth]{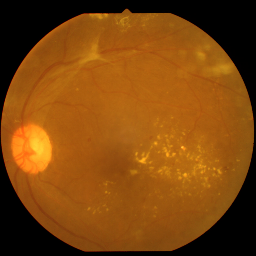}} 
    \subfloat[\small{DME}]{\includegraphics[width=0.18\textwidth]{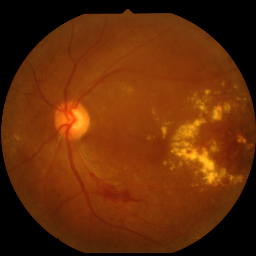}}
    \subfloat[\small{DME}]{\includegraphics[width=0.18\textwidth]{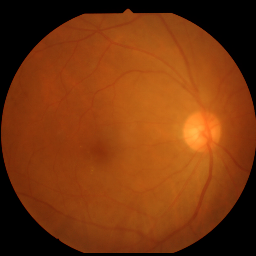}}
    \caption{Sample color fundus photos (CFPs) with and without hard exudates (yellow lesions) and with and without diabetic macular edema (DME) as defined by the more accurate optical coherence tomography (OCT) measurement.}
    \label{fig:dme_image_samples}
\end{figure}

Previous works~\cite{gulshan2016development,krause2018grader} show that a CNN taking CFPs as input can be trained to predict the DME labels derived from human experts grading CFPs. We'll %
refer to such CNNs 
as HE-labels models. %
However, more recent work ~\cite{varadarajan2018predicting} shows that if we train the same CNN architectures using labels derived from human experts grading OCTs, the model (AUC:0.85) can significantly outperform the HE-labels model (AUC:0.74). We'll call these as OCT-labels models.%

\textbf{Key Question} we hope to answer is: What signal is the OCT-labels model capturing to produce such a remarkable increase in AUC, that has not been observed by human experts (and equivalently the HE-labels model)? 
Note that the only difference here is the source of the label (the architecture, and training fundus photos are the same in both cases).

\textbf{DME dataset.} The DME dataset
 used in our experiments come from \cite{varadarajan2018predicting}. %
The dataset consisted of 7072 paired CFP and OCT images taken at the same time. The images were labeled by 2 medical doctors %
(details in \cite{varadarajan2018predicting}). Only the CFPs and the labels derived from the OCTs (and not the OCT images themselves) are used in training the CNN. We used the same datasplits as in \cite{varadarajan2018predicting} with 6039 images for training and validation, and 1033 images for test. %

\textbf{Annotations.}
For the ablation studies, we obtained expert annotations of two known ``landmarks" in CFPs, namely optic disc and fovea. Optic disc is the bright ellpitical region where the optic nerve leaves the retina. 
Fovea is the (dark) area with the largest concentration of cones photoreceptors.
Optic disc was labelled using an ellipse (four parameters) and fovea with a point (2 parameters).

\section{Application}\label{sec:experiments}
As a first step we reproduce results from ~\cite{varadarajan2018predicting} building a CNN model $M$ based on inception-v3 (initializing weights from imagenet) and train on the DME dataset. This model achieves an AUC of 0.847. We also train a multi-task version of this model (which predicts other metadata labels) as described in ~\cite{varadarajan2018predicting} which gets an AUC of 0.89. Next we perform input ablations based on annotated landmarks. 
\begin{figure}%
    \centering
    \subfloat[Models trained on crops centered on the optic disc.]{{\includegraphics[width=0.34\textwidth]{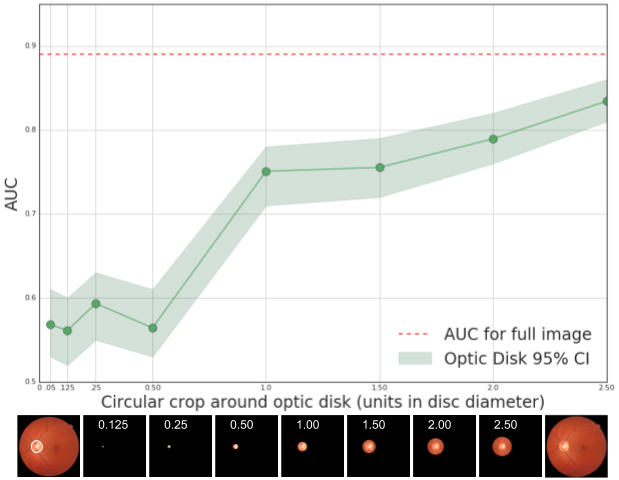} }} %
   \quad
    \subfloat[Models trained on crops centered on the fovea.]{{\includegraphics[width=0.34\textwidth]{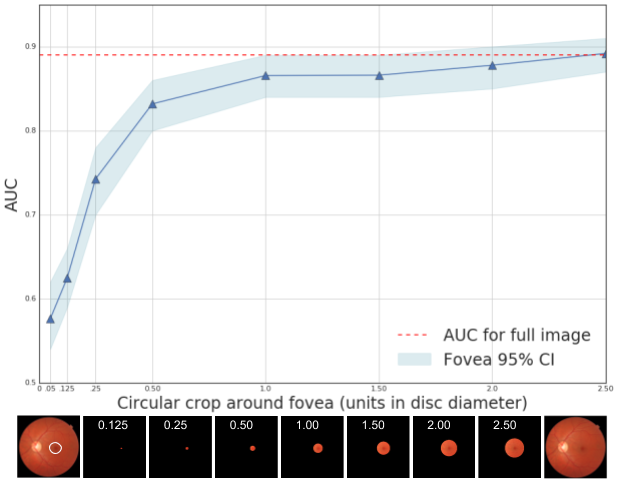} }}%
    \caption{[Input ablations based on \cite{varadarajan2018predicting}] Performance (AUC) of models trained and evaluated on circular crops extracted by centering either on the (a) optic-disc, or the (b) fovea. Bottom panel of images show examples of cropped regions as the diameter of extraction is increased. The horizontal red dashed line is the performance of the model trained on images with no cropping (i.e the OCT-labels model).}%
    \label{fig:ablation}%
\end{figure}

\noindent\textbf{1. Input ablations indicate fovea is important for prediction.}
We extract circular crops of different radii ($0.25$ to $5$), measured relative to the optic disc diameter, around both landmarks (replacing other pixels with background color). Examples of cropped images are shown in %
Fig~\ref{fig:ablation} below each plot. 
This creates 16 ``cropped" versions of the dataset. We train a model %
on each cropped version. We then compare the performance of these models against that of the model with no cropping (i.e the OCT-labels model) using the %
AUC metric. %
As noted in \cite{varadarajan2018predicting} and the plots in Fig.~\ref{fig:ablation}b we can observe that the model gets most of its information from the region surrounding the fovea (i.e the macula). 
This %
serves to reassure that the model is looking at the right region (macula). %

\begin{figure}%
    \centering
    \includegraphics[width=\textwidth]{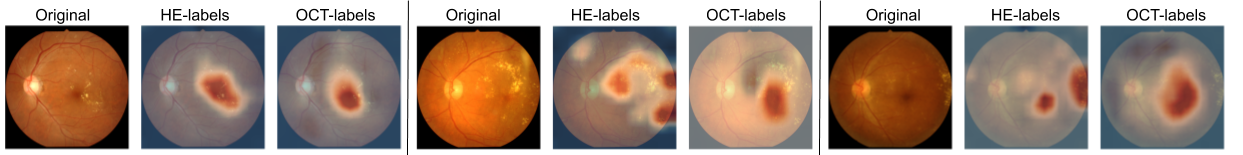}
    \caption{\small{Comparing visual saliency maps from applying GradCAM on the HE-labels and OCT-labels models on 3 sample images with DME from the tune set. Each example shows the original CFP (left), saliency from HE-labels model overlaid on the CFP (middle), and saliency from OCT-labels model overlaid on the CFP (right). OCT-labels model centers around the fovea (the dark region near the center of the image), and the HE-labels model focuses on hard-exudates (the yellow lesions in the image). [Best viewed electronically in high resolution]}}%
    \label{fig:he_vs_oct}%
\end{figure}

\noindent\textbf{2. Saliency maps show where HE-labels and OCT-labels models differ.}
We applied GradCAM~\cite{selvaraju2017grad} on the HE-labels and OCT-labels models to visually see the differences between the two. Qualitative examples (as shown in Fig.~\ref{fig:he_vs_oct}) indicate that, the OCT-labels model focuses on the macula while the HE-labels model focuses on the hard-exudate (yellow lesions) locations. Recall that for the HE-labels model, the doctors use the presence of hard-exudates to determine testing for DME.
The advantage with visual saliency methods are that they offer qualitative explanations without the need for landmarks. Our next step is to understand what the OCT-labels model picks up from the macula. %

\begin{figure}
    \centering
     \subfloat[$f(y)$ model]{\raisebox{-.45\height}{\includegraphics[width=0.26\textwidth,height=2.5cm]{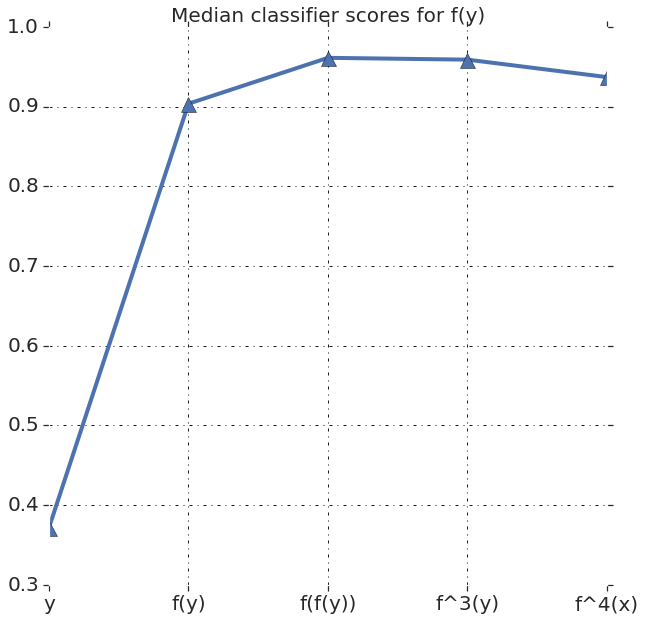} }} %
     \subfloat[$g(x)$ model]{\raisebox{-.45\height}{\includegraphics[width=0.26\textwidth,height=2.5cm]{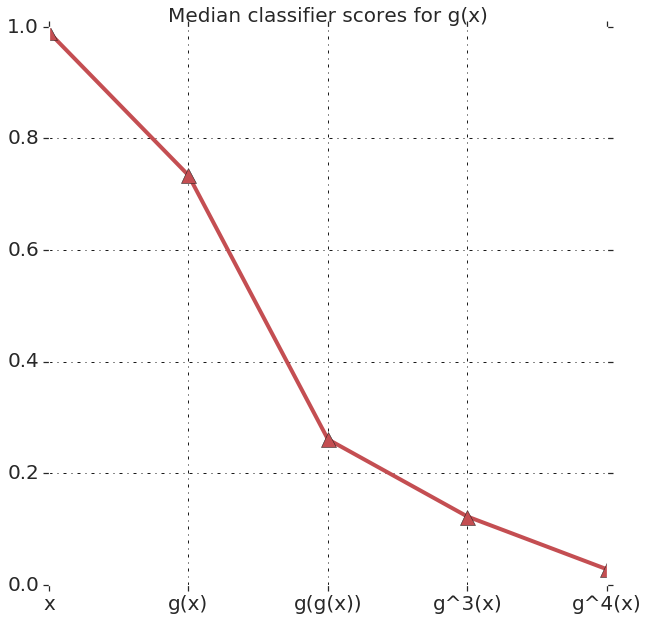} }}
        \subfloat[AUCs for $(g^n(x), f^n(y))$]{
        \begin{tabular}{|c c|} 
 \hline
 \small{Input} &  \small{AUC [95\% CI range]} \\
 \hline
 $(x, y)$ & 0.804 [0.746 - 0.862] \\
 $g(x), f(y)$ & 0.374 [0.294 - 0.450]  \\
 $g^2(x), f^2(y)$ & 0.180 [0.139 - 0.241]  \\
 $g^3(x), f^3(y)$ & 0.130 [0.087 - 0.186]  \\
 $g^4(x), f^4(y)$ & 0.106 [0.060 - 0.156]  \\ [0.5ex] 
 \hline
\end{tabular}
}
    \captionlistentry[table]{A table beside a figure}
    \captionsetup{labelformat=andtable}
    \caption{\small{CycleGAN validation using prediction model $M$ on a random subset (200 images) of the tune set.
    Plots (a) and (b) show median prediction scores on successive applications of $f$ (on DME images) and $g$ (on no-DME images). Table (c) shows AUCs for all the transformed images $(g^n(x), f^n(y))$. 
    These show that the CycleGAN is able to successfully fool prediction model $M$ into thinking images are from the opposite class, and continues to improve with successive applications of $g$ and $f$.}}
    \label{fig:cyclegan_plots}
\end{figure}

\noindent\textbf{3. Modified CycleGANs learn to successfully translate between classes.}
We train a CycleGAN model as described in Sec.~\ref{sec:methods} on the training images with DME ($X$) and no-DME ($Y$). 
We evaluated AUC metric on the transformed images as measured by the independent prediction model $M$ on a random subset of the evaluation set (200 images with near 50-50\% split). Results in Table 1 %
show that our model is able to successfully fool (to an extent) an independently trained supervised classifier on the same task. Specifically, Fig.~\ref{fig:cyclegan_plots} (a, b) and Table 1 (c) show that with successive applications $f$ is able to make $M$ classify translated DME images as no-DME, and $g$ is able to make $M$ classify no-DME images as having DME. %

Note that there is no explicit link between classifier $M$ and the CycleGAN model. In general, we can imagine scenarios where the two may not agree, and hence the analysis in Tab.1(c) is a critical component of the evaluation. The drop in classification model's AUC on repeated CycleGAN application (shown in Fig.~\ref{fig:cyclegan_plots}) provides quantitative evidence that both models are learning similar high level features that distinguish the two classes in the dataset manifold.

\begin{figure}
\captionsetup[subfigure]{labelformat=empty}
    \centering
     \subfloat[\small{$f$: DME to No-DME transformations.}]{\includegraphics[width=0.45\textwidth]{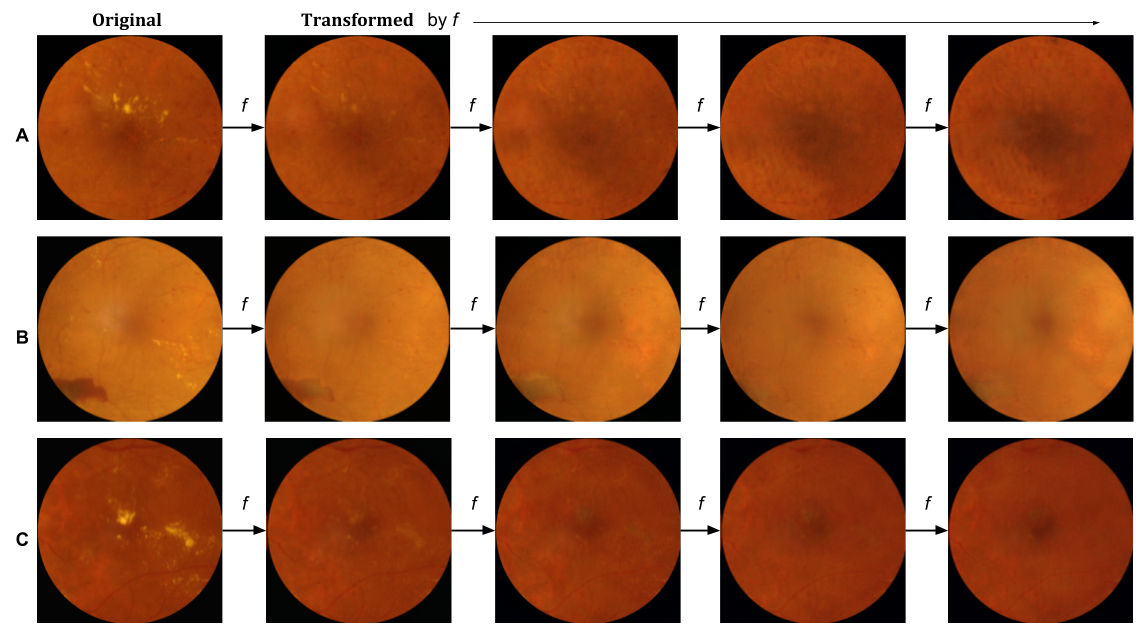}} \quad \qquad
     \subfloat[\small{$g$: No-DME to DME transformations.}]{\includegraphics[width=0.45\textwidth]{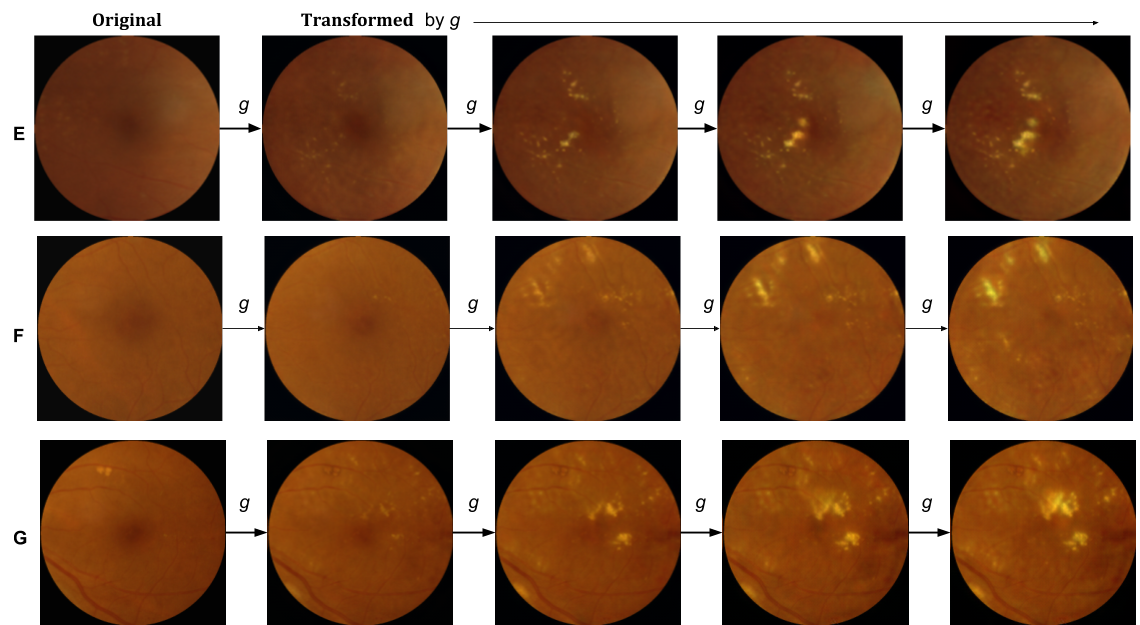}}
     \caption{\small{A selected sample of images showing successive application of $f$ (left) and $g$ (right). Notice that $f$ not only
     removes hard exudates (yellow lesions) but also tends to darken the region around fovea. We believe this accentuates the foveal dip. Notice how $g$ adds both the hard exudates and lightens the macula region. %
     }}
    \label{fig:cyclegan_qualitative}
\end{figure}

\noindent\textbf{4. Amplification highlights hard exudates and brightening in macula.}
With confidence in our CycleGAN transformations we next qualitatively analyze the changes introduced by the model in Figure \ref{fig:cyclegan_qualitative}.
With successive application of $g$ and $f$ two changes are observed: \begin{itemize}[noitemsep,topsep=0pt]
    \item hard exudates are either added (for $g$) or removed (for $f$) consistently.
    \item fovea region is brightened (for $g$) or darkened (for $f$) consistently.
\end{itemize}

\textbf{Hard exudates} Fig \ref{fig:cyclegan_qualitative} shows $g$ (right images) adds hard exudates (yellow lesions) to images and $f$ (left images) generally diminishes or removes hard exudates completely. This is in line with our expectations since hard exudates are correlated (although not perfectly) with DME diagnosis.

\textbf{Fovea brightening.}
The other is a very subtle difference only visible with successive application and with the images best viewed as a gif or video. For transformations in $f$ (left images), in each subsequent image the fovea region gets darker than the surroundings. And in $g$ the fovea region gets progressively lighter. %
Although this change is amplified (visible to the human eye) only with successive applications of functions $f$ and $g$, a single application appears enough to convince classifier $M$ (Table 1) that the classes have changed. This highlights one of the key features of our explanation method %
i.e we can amplify and enhance subtle changes to make them more human interpretable.

\noindent\textbf{5. Converting CycleGAN hypotheses to hand-engineered features.}\label{sec:hand_engineered}
Following the insights from amplification, we engineered a rather simple set of features to objectively evaluate whether the hypotheses given by the model is valid. We took 10 concentric circles (5mm apart) around the fovea, then computed the mean red, green and blue intensities within these discs to create our features. We train a linear SVM classifier and an MLP with 1 hidden layer on these features.

Results in Table \ref{tab:hand-eng} show that the presence(/absence) of hard exudates is complementary to our hand engineered features and boosts the overall AUC. This hand-engineered model used nothing but the average color intensities around the fovea. %
Our hand engineered features along with the presence(/absence) of hard exudates gets very close to explaining the totality of AUC of the CNN model $M$. This builds further confidence in our explanation method.
\begin{table}
\small
\setlength{\tabcolsep}{5pt}
\begin{center}
\begin{tabular}{lc|c}
\toprule
Features & \textbf{SVM} (AUC) & \textbf{MLP} (AUC) \\
\midrule
Hand-engineered features alone  & 72.4 $\pm$ 0.0 & 76.3 $\pm$ 0.3 \\
Presence of hard exudates alone & 74.1 $\pm$ 0.0 & 74.1 $\pm$ 0.0  \\
Hand-engineered features & & \\
 $\quad$ + hard exudates' presence & {\textbf{81.4 $\pm$ 0.0}} & {\textbf{82.2 $\pm$ 0.2}} \\
 \midrule
 $M$(raw pixels single task on cropped image) & \multicolumn{2}{c}{{CNN: 84.7}} \\
\bottomrule
\end{tabular}
\vspace{4pt}
\caption[hand-eng-feats]{\small{
Performance of SVM and MLP classifiers on features hypothesized by our image translation model on the tune set: either hand-engineered features alone, presence of hard-exudates alone, or combination of the two. AUC and standard deviation of 10 runs reported in percentage (higher is better).}
}\label{tab:hand-eng}
\end{center}
\end{table}

\section{Conclusions}
In this paper, we propose a framework to convert novel predictions fueled by deep learning into direct insights for scientific discovery. We believe our image translation based explanation methods are a much more generic tool than ablation or image saliency based methods. Our method goes beyond spatial support to reveal the nature of change between classes which is evaluated objectively. %
We successfully applied this to explain the difference between 2 diabetic macular edema classifiers trained on different sources of labels.  %
Our method is able to provide insights previously unknown to the scientific community.

{\small
\bibliographystyle{splncs04}  
\bibliography{refs_miccai}
}

\section*{Supplementary Material}

\subsection*{Additional Examples of Transformed Images}
\begin{figure}[h]
\captionsetup[subfigure]{labelformat=empty}
    \centering
     \subfloat{\includegraphics[width=\textwidth]{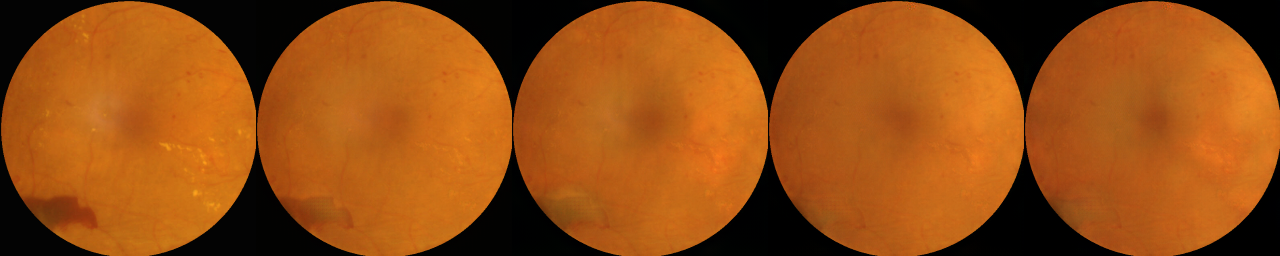}} \\
     \vspace{-1\baselineskip}
     \subfloat{\includegraphics[width=\textwidth]{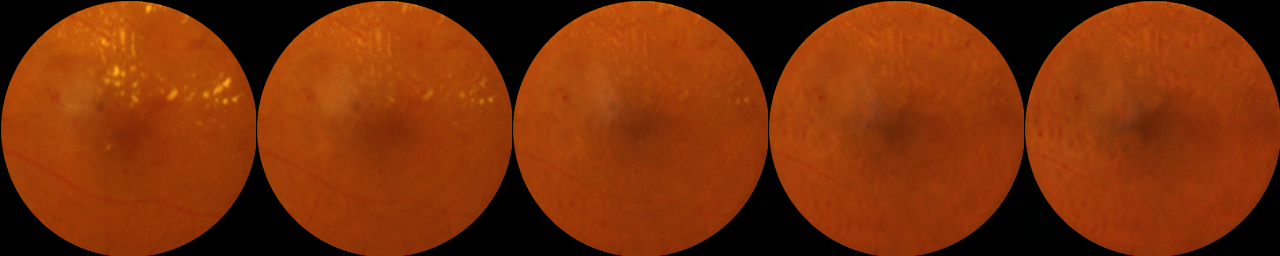}} \\
     \vspace{-1\baselineskip}
     \subfloat{\includegraphics[width=\textwidth]{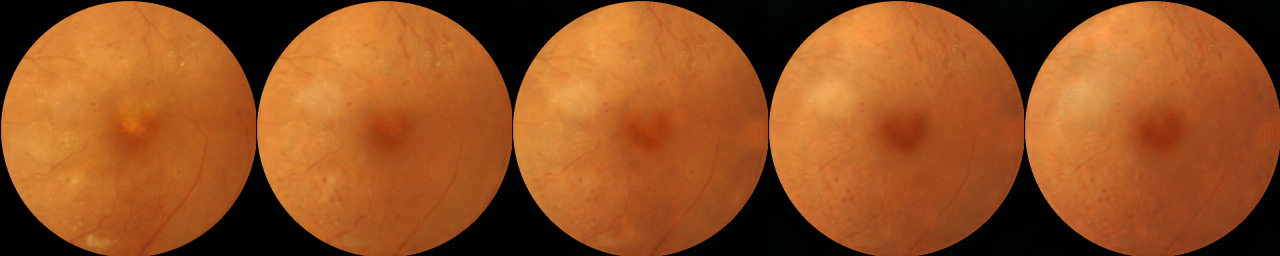}} \\
     \vspace{-1\baselineskip}
     \subfloat{\includegraphics[width=\textwidth]{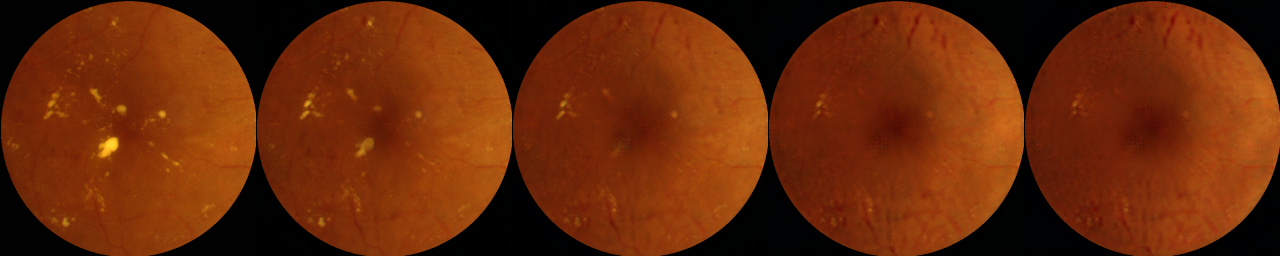}} \\
     \vspace{-1\baselineskip}
     \subfloat{\includegraphics[width=\textwidth]{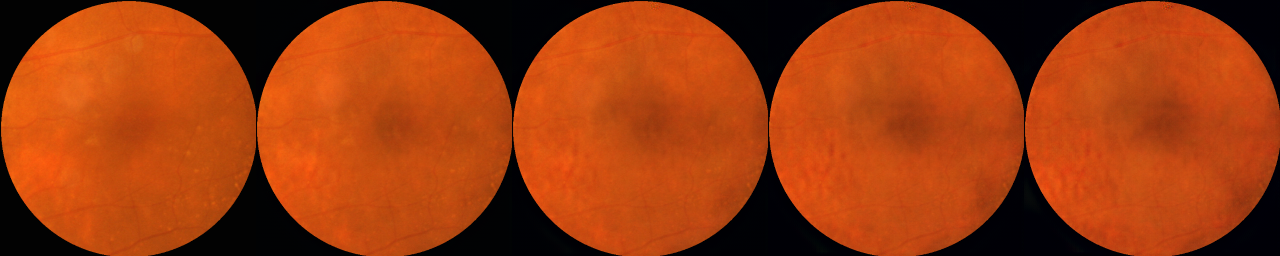}} 
     \caption{\small{A random sample of images showing successive application of $f$. Each column shows successive applications of $f$ from left to right on 5 different input (row) examples. Notice that the $f(\cdot)$ model not only
     removes hard exudates (yellow lesions) but also tends to darken the region around fovea. We believe this accentuates the foveal dip.
     }}
    \label{fig:cyclegan_qualitative_f}
\end{figure}

\begin{figure}[h]
\captionsetup[subfigure]{labelformat=empty}
    \centering
     \subfloat{\includegraphics[width=\textwidth]{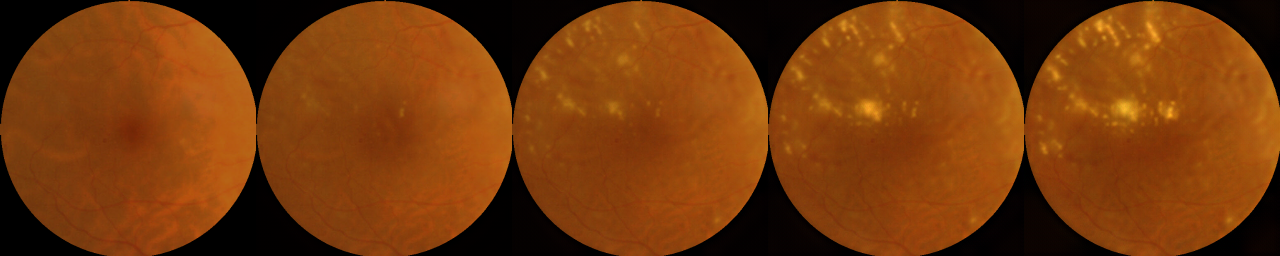}} \\
     \vspace{-1\baselineskip}
     \subfloat{\includegraphics[width=\textwidth]{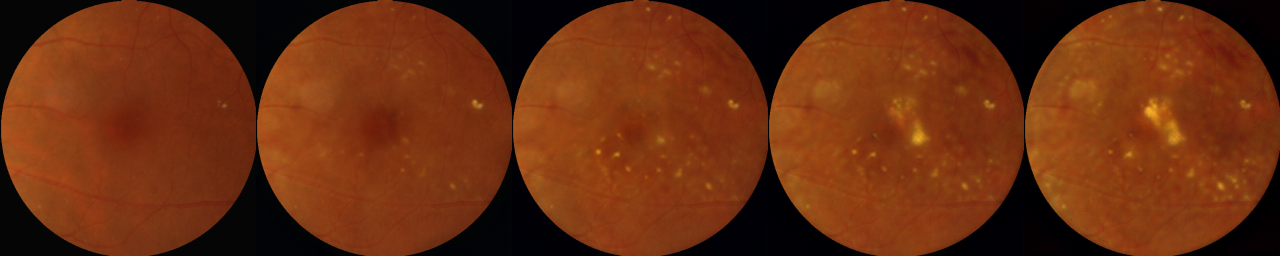}} \\
     \vspace{-1\baselineskip}
     \subfloat{\includegraphics[width=\textwidth]{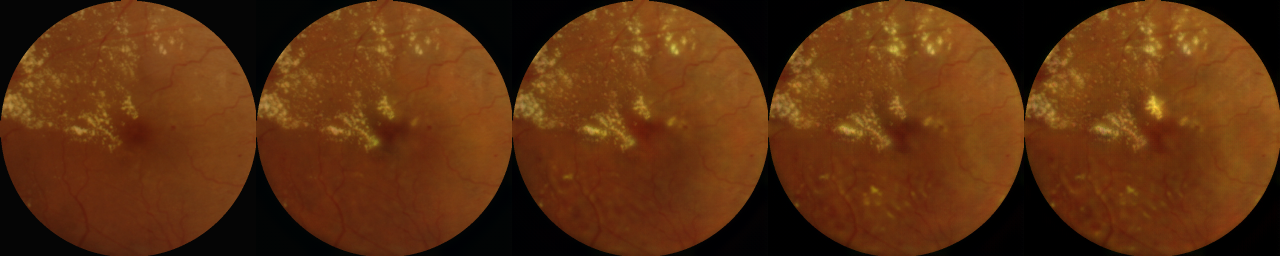}} \\
     \vspace{-1\baselineskip}
     \subfloat{\includegraphics[width=\textwidth]{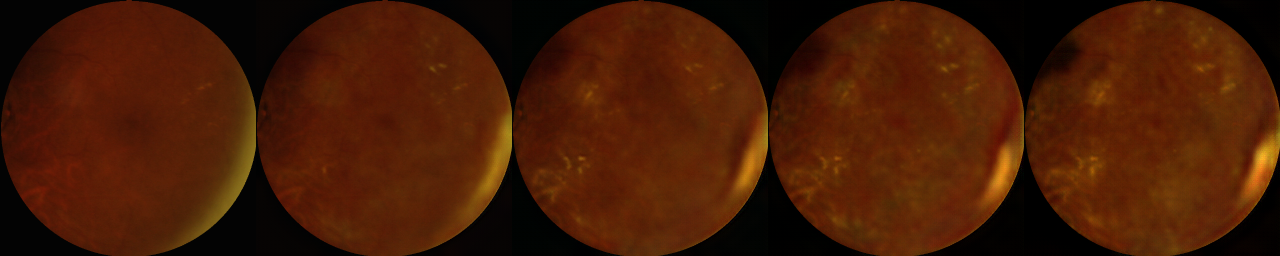}} \\
     \vspace{-1\baselineskip}
     \subfloat{\includegraphics[width=\textwidth]{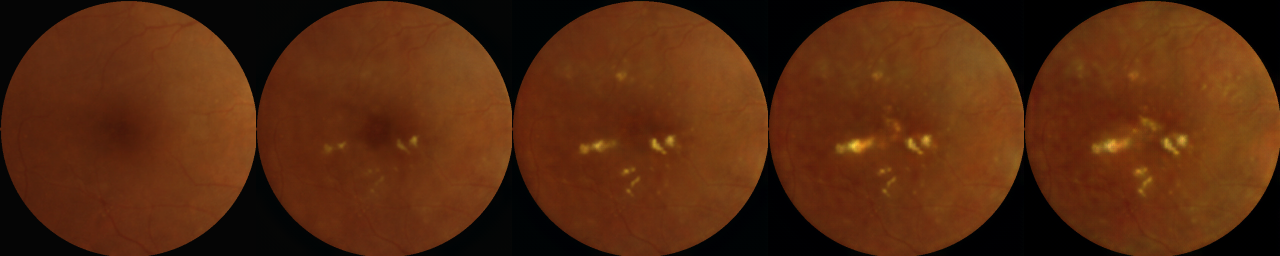}}
     \caption{\small{A random sample of images showing successive application of $g$. Each column shows successive applications of $g$ from left to right on 5 different input (row) examples. Notice how the $g(\cdot)$ model adds both the hard exudates and lightens the macula region. 
     }}
    \label{fig:cyclegan_qualitative_g}
\end{figure}

Figs.~\ref{fig:cyclegan_qualitative_f} and ~\ref{fig:cyclegan_qualitative_g} presents additional examples of transformations generated by the CycleGAN's $f(y)$ and $g(x)$ models respectively.

\textbf{Physical interpretation.} 
The lightening of the fovea can be shown to have a direct physical interpretation. The foveal dip \cite{riordan2011vaughan} exists on the retinal surface around the fovea where most of the visual acuity exists. DME is caused by a swelling of this region and hence affects human vision. We believe the model is able to exploit the minute differences in reflection of light on the retinal surface near fovea to determine whether the person has DME or not. This needs to be calibrated rather precisely, which can be expected from a machine learned model on thousands of images but not humans. We believe this is also the reason why retinal specialists have been unable to capture such features. On the whole, the interpretation that the change in foveal dip causes the light to reflect differently to explain ``\textit{why}" the phenomenon occurs needs to be validated more extensively with paired OCT and fundus images.

\end{document}